\documentclass[runningheads]{llncs}
\usepackage{placeins}

\usepackage{graphicx}
\usepackage{caption} 
\begin{document}
\title{An implementation of the ``Guess who?" game using CLIP}
\titlerunning{An implementation of the ``Guess who" game using CLIP}
%
\author{Arnau Martí Sarri\inst{1}\orcidID{0000-0001-8886-2139} \and \\
Victor Rodriguez-Fernandez \inst{2}\orcidID{0000-0002-8589-6621}}
\authorrunning{A. Martí and V. Rodriguez-Fernandez.}
%
\institute{Valencian International University, Calle Pintor Sorolla 21, 46002 Valencia, Spain \\
Dimai S.L., Camí de la Font Calda 10, 08270 Navarcles, Spain \\
\email{a.marti@dimaisl.com} \and
School of Computer Systems Engineering, Universidad Polit\'ecnica de Madrid, Calle de Alan Turing, 28038 Madrid, Spain\\
\email{victor.rfernandez@upm.es}}
\maketitle              
\begin{abstract}
CLIP (Contrastive Language-Image Pretraining) is an efficient method for learning computer vision tasks from natural language supervision that has powered a recent breakthrough in deep learning due to its zero-shot transfer capabilities. By training from image-text pairs available on the internet, the CLIP model transfers non-trivially to most tasks without the need for any data set specific training. In this work, we use CLIP to implement the engine of the popular game ``Guess who?", so that the player interacts with the game using natural language prompts and CLIP automatically decides whether an image in the game board fulfills that prompt or not. We study the performance of this approach by benchmarking on different ways of prompting the questions to CLIP, and show the limitations of its zero-shot capabilites.

\keywords{CLIP \and Guess who \and Zero-shot learning \and Language-image models.}
\end{abstract}
\section{Introduction}
The ability to learn at the same time from different data modalities (image, audio, text, tabular data...) is a trending topic in the field of machine learning in general, and deep learning in particular, with many domains of application such as self-driving cars, healthcare and the Internet of Things \cite{Gao2020ASO}. Among all the possibilities that multimodal deep learning provides, one of the most interesting ones is the connection of text and images in the same model.

This concept brings into play challenging tasks in the areas of computer vision and natural language processing, such as multimodal image and text classification \cite{8960960}, image captioning \cite{cornia2019eshedmemory}, and visual-language robot navigation \cite{tan2019earning}. All of these tasks have the core idea of learning visual perception from supervision contained in text, or vice-versa.

In January 2021, the company OpenAI made a great milestone in the field of language-image models with the presentation of CLIP (Contrastive Language–Image Pre-training) \cite{radford2021earning} \footnote{The paper was accompanied with a blog post publication: \url{https://openai.com/blog/clip/}}. CLIP is a deep neural network designed to perform zero-shot image classification, i.e., to generalize flawlessly to unseen image classification tasks in which the data and the labels can be different each time. The way CLIP does so is by training on a wide variety of (image, text) pairs that are abundantly available on the internet, instructing the model to predict the most relevant text snippet, given an image. Since the code and weights of CLIP were publicly released on GitHub \footnote{\url{https://github.com/openai/CLIP}}, many researchers have explored its zero-shot capabilities in different areas such as art classification \cite{Conde2021CLIPArtCP}, video-text retrieval \cite{Fang2021CLIP2VideoMV} or text-to-image generation \cite{ramesh2021eroshot}.

In this work, we present an application of the zero-shot classification capabilities of CLIP in the popular game ``Guess who?", in which the player asks yes/no questions to describe people in a game board and try to guess who the selected person is. Each time the player makes a new question, CLIP will analyse the images in the game board and decide automatically which images fulfill it. Although this could be also tackled with a multi-label image binary classification model with a fixed set of labels, the power behind using CLIP relies on the use of natural language to interact with the model, which gives freedom to the player to ask any question and tests CLIP's zero-shot generalization capabilities. We release our code in a public Github repository \footnote{\url{https://github.com/ArnauDIMAI/CLIP-GuessWho}}. 

In summary, the contributions of this paper are:
\begin{itemize}
    \item The implementation of the game engine based on AI, which allows for the use of any set of images instead of having a fixed board. To the best of our knowledge, there is no other version of the game ``Guess who?", that uses an AI in a similar way.
    \item The use of CLIP as a zero-shot classifier based on textual prompts, which allows the player to interact with the game through natural language.
\end{itemize}

The rest of the paper is structured as follows: in Section \ref{sec:backgrounds} we give some background on CLIP, in Section \ref{sec:game_description} we present a description of the game and how CLIP is integrated in it, not as a player but as the engine of the game. Then, in Section \ref{sec:experiments} we show some experiments on how changing the way the game prompts CLIP about the characteristic of a person affects its classification performance, and finally, in Section \ref{sec:conclusions} we outline the conclusions and provide future lines of research in this topic.

\section{Backgrounds on CLIP}
\label{sec:backgrounds}

CLIP (Contrastive Language-Image Pre-training), by OpenAI, is based on a large amount of work in zero-shot transfer, natural language supervision and multi-modal learning, and shows that scaling a simple pre-training task is sufficient to achieve competitive zero-sample performance on various image classification data sets. CLIP uses a large number of available supervision sources: the text paired with images found on the Internet. This data is used to create the following agent training task for CLIP: Given an image, predict which of a set of 32,768 randomly sampled text fragments is actually paired with it in the data set. This is achieved by combining a text encoder, built as a Transformer \cite{vaswani2017ttention}, and an image encoder, built as a Vision Transformer \cite{dosovitskiy2020n}, under a contrastive objective that connects them \cite{zhang2020contrastive}. To the best of our knowledge, there is no other publicly available pretrained model with the scale of CLIP that connects text and image data.

Once pre-trained, CLIP can then be applied to nearly arbitrary visual classification tasks. For instance, if the task of a data set is classifying photos of dogs vs cats, we will check, for each image, whether CLIP predicts that the caption “a picture of a dog” is more likely to be paired with it than ``a picture of a cat". In case the task does not have a fixed set of labels, one can still use CLIP for classification by specifying a text description of a target attribute and a corresponding neutral class. For example, when manipulating images of faces, the target attribute might be specified as ``a picture of a blonde person", in which case the corresponding neutral prompt might be ``a picture of a person". CLIP's zero-shot classifiers can be sensitive to wording or phrasing, and sometimes require trial and error ``prompt engineering" to perform well \cite{radford2021earning}.

\section{Game description}
\label{sec:game_description}

\begin{figure*}[ht!]
\noindent\makebox[\textwidth]{\frame{\includegraphics[scale=0.17]{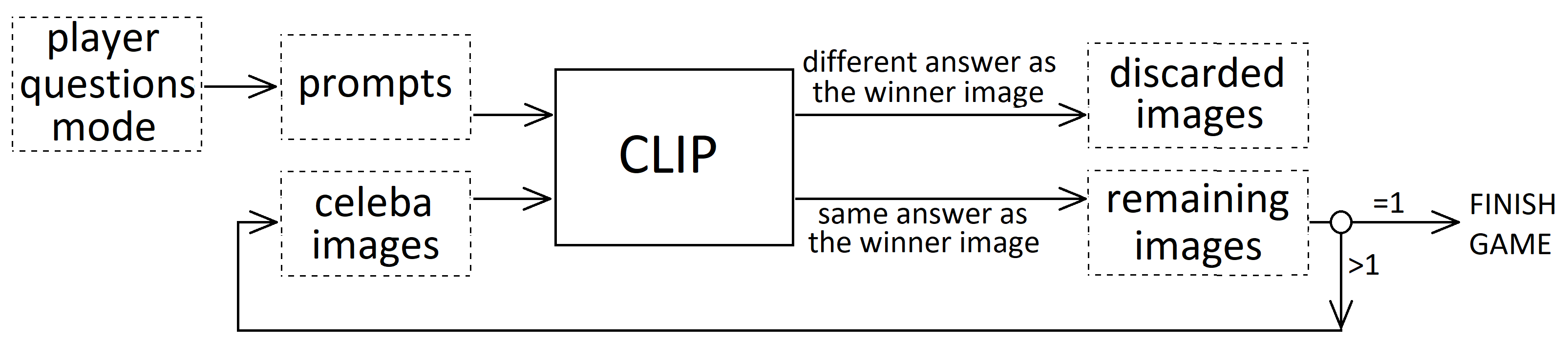}}}
\caption{Diagram of how CLIP is integrated in the game.
   \label{fig:CLIPdiagram}}
\end{figure*}

The aim of the game, as in the original one, is to find a specific image from a group of different images of a person's face. To discover the image, the player must ask questions that can be answered with a binary response, such as ``Yes and No". After every question made by the player, the images that don't share the same answer that the winning one are discarded automatically. The answer to the player's questions, and thus, the process of discarding the images will be established by CLIP (See Fig. \ref{fig:CLIPdiagram}). When all the images but one have been discarded, the game is over.

\begin{figure*}[ht!]
\noindent\makebox[\textwidth]{\frame{\includegraphics[scale=0.7]{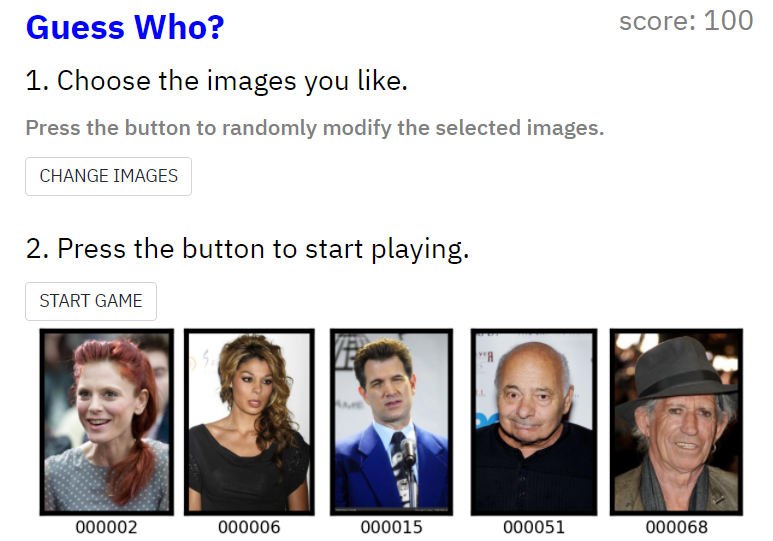}}}
\caption{Screen of the game which allows the user selecting the images to play.
   \label{fig:select_images}}
\end{figure*}

The first step of the game is to select the images to play (See Fig. \ref{fig:select_images}). The player can press a button to randomly change the used images, which are taken from the \textit{CelebA} (CelebFaces Attributes) data set \cite{liu2015deep}. This data set contains 202,599 face images of the size 178×218 from 10,177 celebrities, each annotated with 40 binary labels indicating facial attributes like hair color, gender and age.

\subsection{Questions}

The game will allow the player to ask the questions in 4 different ways:

\begin{enumerate}

\begin{figure*}[ht!]
\noindent\makebox[\textwidth]{\frame{\includegraphics[scale=0.7]{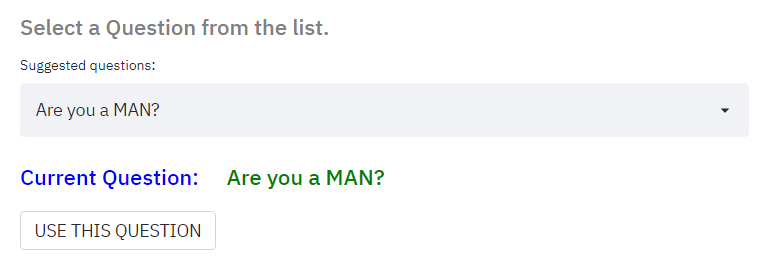}}}
\caption{Game screen that allows the user to ask a default question.
   \label{fig:ask_question}}
\end{figure*}

\item Asking a question from a list (See Fig. \ref{fig:ask_question}). A drop-down list allows the player to select the question to be asked from a group of pre-set questions, taken from the set of binary labels of the Celeba data set. Under the hood, each question is translated into textual prompts for the CLIP model to allow for the binary classification based on that question. When they are passed to CLIP along with an image, the model responds by giving a greater value to the prompt that is most related to the image.

\begin{figure*}[ht!]
\noindent\makebox[\textwidth]{\frame{\includegraphics[scale=0.7]{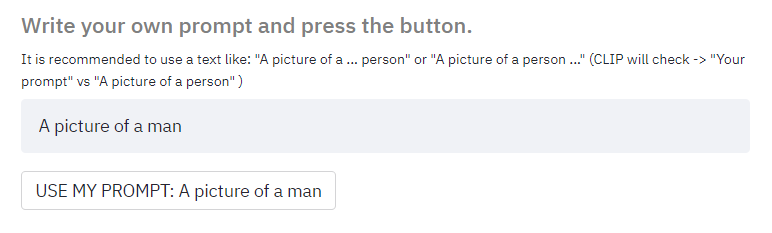}}}
\caption{Game screen that allows the user to create his own prompt using 1 text input.
   \label{fig:1_query}}
\end{figure*}

\item Use one prompt (See Fig. \ref{fig:1_query}). This option is used to allow the player introducing a textual prompt for CLIP with his/her own words. The player text will be then confronted with the neutral prompt, ``A picture of a person", and the pair of prompts will be passed to CLIP as in the previous case.

\begin{figure*}[ht!]
\noindent\makebox[\textwidth]{\frame{\includegraphics[scale=0.7]{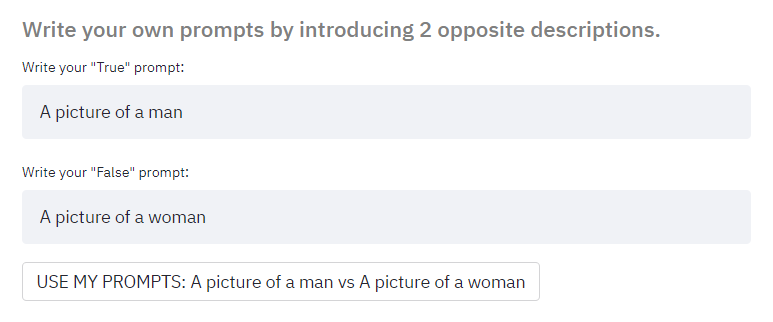}}}
\caption{Game screen that allows the user to create his own prompt using 2 text inputs.
   \label{fig:2_querys}}
\end{figure*}

\item Use two prompts (See Fig. \ref{fig:2_querys}). In this case two text input are used to allow the player write two sentences. The player must use two opposite sentences, that is, with an opposite meaning.

\begin{figure*}[ht!]
\noindent\makebox[\textwidth]{\frame{\includegraphics[scale=0.7]{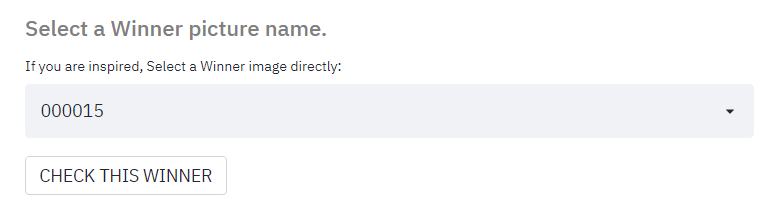}}}
\caption{Game screen that allows the user to select the winner image directly.
   \label{fig:select_winner}}
\end{figure*}

\item Select a winner (See Fig. \ref{fig:select_winner}). This option does not use the CLIP model to make decisions, the player can simply choose one of the images as the winner and if the player hits the winning image, the game is over.

\end{enumerate}

\subsection{Punctuation}

To motivate the players in finding the winning image with the minimum number of questions, a scoring system is established so that it  begins with a certain number of points (100 in the example), and decreases with each asked question. The score is decreased by subtracting the number of remaining images after each question. Furthermore, there are two extra penalties. The first is applied when the player uses the option ``Select a winner". This penalty depends on the number of remaining images, so that the fewer images are left, the bigger will be the penalty. Finally, the score is also decreased by two extra points if, after the player makes a question, no image can be discarded.

\begin{figure*}[ht!]
\noindent\makebox[\textwidth]{\frame{\includegraphics[scale=0.7]{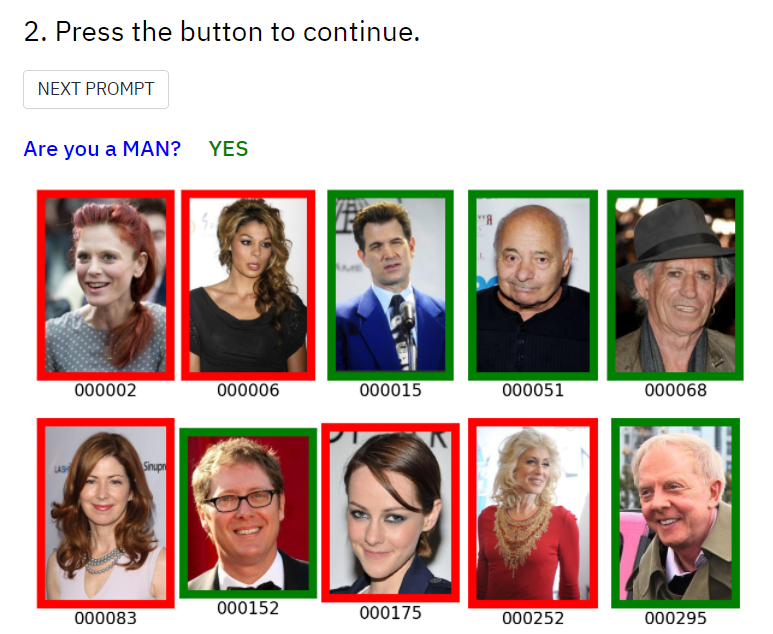}}}
\caption{Game screen showing CLIP answer after a question.
   \label{fig:show_CLIP_results}}
\end{figure*}

The ``Guess Who?" game has a handicap when it uses real images, because it is necessary to always ensure that the same criteria are applied when the images are discarded. The original game uses images with characters that present simple and limited features like a short set of different types of hair colors, what makes it very easy to answer true or false when a user asks for a specific hair color. However, with real images it is possible to doubt about if a person is blond haired or brown haired, for example, and it is necessary to apply a method which ensures that the winning image is not discarded by mistake. To solve this problem, CLIP is used  to discard the images that do not coincide with the winner image after each prompt. In this way, when the user asks a question, CLIP is used to classify the images in two groups: the set of images that continue because they have the same prediction than the winning image, and the discarded set that has the opposite prediction. Fig. \ref{fig:show_CLIP_results} shows the screen that is prompted after calling CLIP on each image in the game board, where the discarded images are highlighted in red and the others in green.

\section{Experiments and prompt analysis}
\label{sec:experiments}

To use CLIP as a zero-shot image classifier in the game, we create a pair of textual prompts for each class to address each player question as its own binary classification problem. Two basic prompting methods are proposed to create the textual descriptions:

\begin{enumerate}
\item Target vs neutral. This method consists in using a standard neutral prompt that fulfills all images, like ``A picture of a person", and another sentence, very similar, which changes only some words and is more specific for the target class, like ``A picture of a person with eyeglasses". In this way, when the additional information is added to the prompt, CLIP should return a bigger value for the second sentence than for the first for an image of a person with eyeglasses. And vice versa, i.e., when the extra information is not related to the image, CLIP should return a smaller likelihood value for this sentence. This is the method used to allow the user to introduce his own prompt.

\item Target vs contrary. This method consists in using two opposite sentences that represent opposite concepts, like ``A picture of a man" and ``A picture of a woman". This method is only implemented in the game for the attributes included in the set of labels of the Celeba data set.

\end{enumerate}


We take advantage of the labeled images from the Celeba data set to validate the performance of the textual prompts introduced to CLIP. The True Positive Rate (TPR), True Negative Rate (TNR), and the accuracy (average of TPR and TNR) are calculated to analyze the results. We used the first 4.000 true images and the first 4.000 false images for each of the 40 binary labels of the data set to calculate these rates. 

\subsection{Results table for the ``Target vs neutral" method}

\begin{table}[]
\begin{tabular}{llccc}
\multicolumn{1}{c}{\textbf{Celeba label}} & \multicolumn{1}{c}{\textbf{Target prompt}} & \textbf{TPR} & \textbf{TNR} & \textbf{Acc} \\
\hline
male                                                  & A picture of a male person                      & 98.14        & 96.13        & 97.11             \\
wearing hat                                           & A picture of a person with hat                  & 97.29        & 83.67        & 89.34             \\
goatee                                                & A picture of a person with goatee               & 91.6         & 77.05        & 82.78             \\
blond hair                                            & A picture of a person with blond hair           & 74.14        & 97.84        & 82.09             \\
bangs                                                 & A picture of a person with bangs                & 88           & 77.71        & 82.05             \\
eyeglasses                                            & A picture of a person with eyeglasses           & 87.45        & 77.59        & 81.78             \\
smiling                                               & A picture of a person who is smiling            & 89.07        & 76.75        & 81.76             \\
bald                                                  & A picture of a bald person                      & 96.63        & 73.86        & 81.58             \\
wearing necktie                                       & A picture of a person with necktie              & 77.91        & 81.36        & 79.54             \\
gray hair                                             & A picture of a person with gray hair            & 83.66        & 74.04        & 78.05             \\
...                                                     & ...                                               & ...            & ...            & ...                 \\
big lips                                              & A picture of a person with big lips             & 64.64        & 51.75        & 53.12             \\
wearing lipstick                                      & A picture of a person with lipstick             & 85.34        & 51.53        & 52.94             \\
pointy nose                                           & A picture of a person with pointy nose          & 52.74        & 52.16        & 52.41             \\
big nose                                              & A picture of a person with big nose             & 57.63        & 51.09        & 51.91             \\
attractive                                            & A picture of an attractive person               & 54.34        & 50.88        & 51.46             \\
rosy cheeks                                           & A picture of a person with rosy cheeks          & 49.36        & 49.76        & 49.65             \\
high cheekbones                                       & A picture of a person with high cheekbones      & 47.33        & 49.23        & 48.8              \\
bags under eyes                                       & A picture of a person with bags under eyes      & 47.7         & 49.05        & 48.66             \\
narrow eyes                                           & A picture of a person with narrow eyes          & 40.67        & 45.36        & 43.8              \\
no beard                                              & A picture of a person with no beard             & 16.93        & 30.02        & 25.09             \\
\hline
\end{tabular}
\caption{``Target vs neutral" prompting method applied on Celeba data set. True Positive Rate, True Negative Rate and Accuracy are shown in percentage.
   \label{table:celeba_results}}
\end{table}

Table \ref{table:celeba_results} shows the top ten and bottom ten results sorted by accuracy, as well as the labels of the Celeba data set and the CLIP textual inputs used. In this experiment, we simply use the literal Celeba labels to create the target prompt, and the neutral prompt is kept as ``A picture of a person". With all this, approximately 25\% of the target prompts obtained an accuracy above 70\%. The 'male' and the 'wearing hat' features obtained remarkable accuracy results, 97\% and 89\% respectively. 

In general, CLIP works sufficiently well when the label represents a physical object (e.g., hat or eyeglasses) or a common expression (e.g., smile), which will arguably be common in the data set in which CLIP has been trained on. The CLIP data set has millions of images and natural text from the Internet, so we must think about how the image descriptions in Internet often look like, in order to engineer good prompts. For example, when talking about a specific person appearing in an image that contains several people, it is common to talk about ``the one who wears a hat" or ``the one who has a goatee", but is unlikely to use a description like ``the one who has narrow eyes" or ``the one who has rosy cheeks".

Another key for CLIP performance lies in the ambiguity of elements or concepts. Some labels present no doubt about whether they are true or false, but some others are susceptible to observer interpretation. For instance, asking if a person wears a hat is a very objective concept that raises no doubt, so practically everyone would respond the same. However, asking if a person has big lips, has a pointy nose, or is attractive, are relative or subjective questions whose answer will depend on the observer. In these cases, it seems that CLIP does not work so well.

Finally, a remarkable result in this experiment is the accuracy obtained with the feature ``no beard", which is related to a negation. In this case, the accuracy is 25\%, but in a binary classification such as this, what CLIP really has reached is the 75 \% of accuracy, if we invert its response. That means that CLIP ignored the ``no" word in the prompt and classified the images of a person with beard with a 75\% of accuracy. That indicates that CLIP probably is not able to deal with the concept of negation. 

\subsection{Results for the ``Target vs contrary" method}

\begin{table}[]
\begin{tabular}{llcccc}
\multicolumn{1}{c}{\textbf{Target prompt}} & \multicolumn{1}{c}{\textbf{Contrary prompt}} & \textbf{TPR} & \textbf{TNR} & \textbf{Acc} & \textbf{Gain} \\
\hline
A picture of a man                               & A picture of a woman                            & 99.39        & 97.72        & 98.54             & +1.43                \\
A picture of a bald person                       & A picture of a haired person                    & 96.08        & 80.43        & 86.65             & +5.07                \\
A picture of a person who             & A picture of a person who            & 80.05        & 89.88        & 84.28             & +2.52                \\
is smiling             & is serious            &         &         &              &   \\
A picture of a person with             & A picture of a person with          & 68.12        & 75.8         & 71.29             & +3.35                \\
pale skin             & tanned skin          &         &          &              &  \\
A picture of a young person                      & A picture of an aged person                     & 65.17        & 78.2         & 69.72             & +5.59                \\
A picture of a person with         & A picture of a person with            & 60.38        & 67.04        & 62.9              & +6.96                \\
straight hair         & wavy hair            &         &         &               &    \\
A picture of an attractive                & A picture of an unattractive             & 50.44        & 50.13        & 50.2              & -1.26      \\         
person                & person             &         &         &        &       \\
\hline
\end{tabular}
\caption{``Target vs contrary" prompting method applied on Celeba data set. True Positive Rate, True Negative Rate, Accuracy and Gain are shown in percentage.
   \label{table:celeba_results2}}
\end{table}

Table \ref{table:celeba_results2} shows some examples of the performance of the second proposed prompting method. The results of the first method, ``target vs neutral" are also shown for easy comparison. This experiment shows how, in general, this method allows to improve the classification results by prompting ``contrary words" to the target attribute, since. Even the ``male" attribute, which obtained a 97 \% of accuracy in the first experiment, can be improved with this method, reaching almost the 1,5\% of accuracy improvement. 

However, it must be remarked that it is difficult to find useful ``contrary words". As an example, the label ``wearing a necktie", ``wearing lipstick" or ``having rosy cheeks" only can be inverted using a negation, what does not work properly with CLIP. In cases where the antonym is very similar to the original word as ``attractive" and ``unattractive", CLIP does not seem to work properly either, but more tests must be done to ensure it.

\section{Conclusions and Future work}
\label{sec:conclusions}

In this work, we present an implementation of the popular game ``Guess Who?", in which the part of the game engine that decides whether an image fulfills the player question or not is made by CLIP, a language-image model that estimates how good a text caption pairs with a given image. To do that, we take each player prompt describing a person attribute and confront it with another prompt that does not represent it. In this way, we can create a binary classifier for each attribute of interest just by getting the maximum likelihood of  CLIP's output to those two prompts. We have tried different prompting methods, such as confronting the target prompt with a neutral one (e.g., ``A picture of a person") or using a prompt that describes the contrary of the target attribute. Experiments have been made with the labelled images of the Celeba data set to analyse the performance of these two prompting methods, showing that, as long as there is a clear contrary word to the target attribute, using the contrary prompt method normally leads to a better zero-shot classification performance.

As future work, we will continue working on ``prompt engineering" due to CLIP-based zero-shot classifiers can be sensitive to wording or phrasing. Moreover, we will deploy the game as a public web application where interested users can play with their own images and give feedback about its performance.

\section*{Acknowledgements}
This work has been partially supported by the company Dimai S.L, the ``Convenio Plurianual" with the Universidad Politécnica de Madrid in the actuation line of ``Programa de Excelencia para el Profesorado Universitario" and by next research projects: FightDIS (PID2020-117263GB-100), IBERIFIER (2020-EU-IA-0252:29374659), and the CIVIC project (BBVA Foundation Grants For Scientific Research Teams SARS-CoV-2 and COVID-19).




%
%
%
\bibliographystyle{splncs04}
\bibliography{mybibliography}
\end{document}